\documentclass{sig-alternate}
\newcommand{\ignore}[1]{}
\usepackage[pass]{geometry}
\usepackage{fancyhdr}
\usepackage[normalem]{ulem}
\usepackage[hyphens]{url}
\usepackage{hyperref}
\usepackage{color}
\usepackage{soul}
\usepackage{multirow}

\fancypagestyle{firstpage}{
  \fancyhf{}
\setlength{\headheight}{50pt}

  \fancyhead[C]{\normalsize{
      \textbf{} }}
  \pagenumbering{arabic}
}

\title{Reliable and Energy Efficient MLC STT-RAM Buffer for CNN Accelerators}
\author{Masoomeh Jasemi $\dagger \star$, Shaahin Hessabi$\star$, and Nader Bagherzadeh$\dagger$\\\\
$\star$Department of Computer Engineering, Sharif University of Technology, Tehran\\
$\dagger$Computer Science Department, University of California, Irvine, Irvine\\
{jasemi@ce.sharif.edu,mjasemi@uci.edu}
{hessabi@sharif.edu}
{nader@uci.edu}
}

\begin{document}

\maketitle
\thispagestyle{firstpage}
\pagestyle{plain}


\begin{abstract}

We propose a lightweight scheme where the formation of a data block is changed in such a way that it can tolerate soft errors significantly better than the baseline. The key insight behind our work is that CNN weights are normalized between -1 and 1 after each convolutional layer, and this leaves one bit unused in half-precision floating-point representation. By taking advantage of the unused bit, we create a backup for the most significant bit to protect it against the soft errors. Also, considering the fact that in MLC STT-RAMs the cost of memory operations (read and write), and reliability of a cell are content-dependent (some patterns take larger current and longer time, while they are more susceptible to soft error), we rearrange the data block to minimize the number of costly bit patterns. Combining these two techniques provides the same level of accuracy compared to an error-free baseline while improving the read and write energy by 9\% and 6\%, respectively.

\end{abstract}

\section{Introduction}
\label{sec:introduction}
Convolutional neural networks (CNNs) are one of the most popular deep learning structures. These models are consisted of many layers with different functionalities to perform some tasks that are usually difficult for the traditional algorithms \cite{CNN1} \cite{CNN3}. CNNs have been around for many years and studied deeply over 25 years after LeNet5 was proposed \cite{LENET5}. However, they became popular when the inventions in computer architectures paved the way for programmable and massive parallel computation needed by these structures. General-purpose Graphical processing Units (GP-GPU) allow the CNN computation to be carried out quickly and easily thanks to the recent progress on programming models and architecture advancements \cite{GPU, LUlaw}. However, the push from CNN designers to create deeper and larger models in one hand, and high power consumption of GPUs on the other hand, motivated the architects to increase the computation capacity of current platforms by suggesting new special-purpose accelerators \cite{Diannao,Eyeriss}. 

The CNN computations are inherently parallel, and they require a large amount of memory due to their large working set size. One solution to overcome the problem of larger memory is employing emerging memory technologies such as STT-RAM or eDRAM. Non-volatile memories provide higher capacity (4X or more) at almost the same area \cite{STT2,STT3}. The other appealing feature of NVMs is called Multi-Level Cell where more than one bit can be stored in any single cell. More clearly, by using a more sophisticated circuit, the resistance spectrum of these resistive memories can be partitioned into more than two regions, and then more than one bit can be stored in a single cell \cite{MLC_1,MLC_2}. This feature is not free and imposes some major challenges. The reliability of MLC STT-RAM is lower than that of SLC, and can be as high as $1.5 \times 10 ^{-2}$ to $2 \times 10 ^{-2}$ \cite{staterestrict}. The lifetime of SLC STT-RAM devices fabricated so far is less than $4 \times 10 ^{15}$ cycles, which is very close to conventional memories. However for MLC STT-RAM, the larger write current exponentially degrades the lifetime \cite{endurance}. So, to benefit from the larger capacity of MLC STT-RAM, the major weaknesses associated with MLC NVMs such as low reliability, high dynamic power consumption and shorter lifetime must be addressed comprehensively.

Fortunately, CNN models are naturally robust to some level of inaccuracy. In other words, the accuracy of prediction will not significantly drop, if the weights slightly change either intentionally by the designer to reduce the space, or by the memory technology substrate which might not be highly reliable \cite{Stochastic,DeepBurning}. However, naively replacing the memory system with a low reliable one may impact the accuracy; as we will show in the later sections. Thus, we are seeking for a larger MLC STT-RAM memory to replace the traditional SRAM memories while maintaining the prediction accuracy. 

In this paper, we propose two simple yet effective schemes to efficiently tolerate the soft errors and also at the same time reduce the energy dissipation. The first scheme which is called \emph{Sign-Bit Protection} utilizes an unused bit in half-precision floating-point representation to duplicate the sign-bit. Based on our experiment sign-bit error is the main contributor to accuracy loss and, thus must be protected separately. We show that protecting the sign bit can be done for free because we can duplicate it in an unused bit. The key behind this scheme is that weights are normalized between -1 and 1 after each convolutional layer, and the second bit in half-precision floating-point representation remains unused. This duplication allows us to change the cell mode from vulnerable MLC mode to safe and reliability-friendly SLC mode. 

Additionally, we propose a data reformation scheme where by manipulating the data, we increase the error resiliency of the system. The key behind the second scheme is that some bit patterns are power-friendly and at the same time they are more robust to soft errors (i.e in a 2-bit MLC-STT, "00" is easier to program and also has higher soft error resiliency \cite{staterestrict}), while other patterns are not very robust. We manipulate the content of the data block by simple operations such as rotation and rounding to increase the number of reliability- and power-friendly patterns and minimize the power-hungry and vulnerable patterns. Combining these two schemes guarantees the accuracy of prediction to be as good as the error-free baseline while increasing the energy efficiency. 

Our experimental results taken from TensorFlow \cite{TensorFlow} platform and SCALE-Sim models \cite{SCALESim} show that our scheme can provide 89\% and 97\%  top-5 accuracy of prediction for ImageNet and VGG16 while reducing the read and write energy dissipation by 9\% and 6\%, respectively. Our scheme needs 2 bits per 16 bits (12.5\%) and 2 bits per 64 bits (3.125\%) of storage overhead for the most energy-efficient and the energy-balanced systems, respectively; while providing the same level of accuracy compared to the baseline.


\section{Background}
In this section, we provide the required backgrounds of CNN accelerators and MLC STT-RAM. 
\label{sec:Background}
\subsection{CNN Accelerator}
\begin{figure*}
  \centering
   \includegraphics[width=5in]{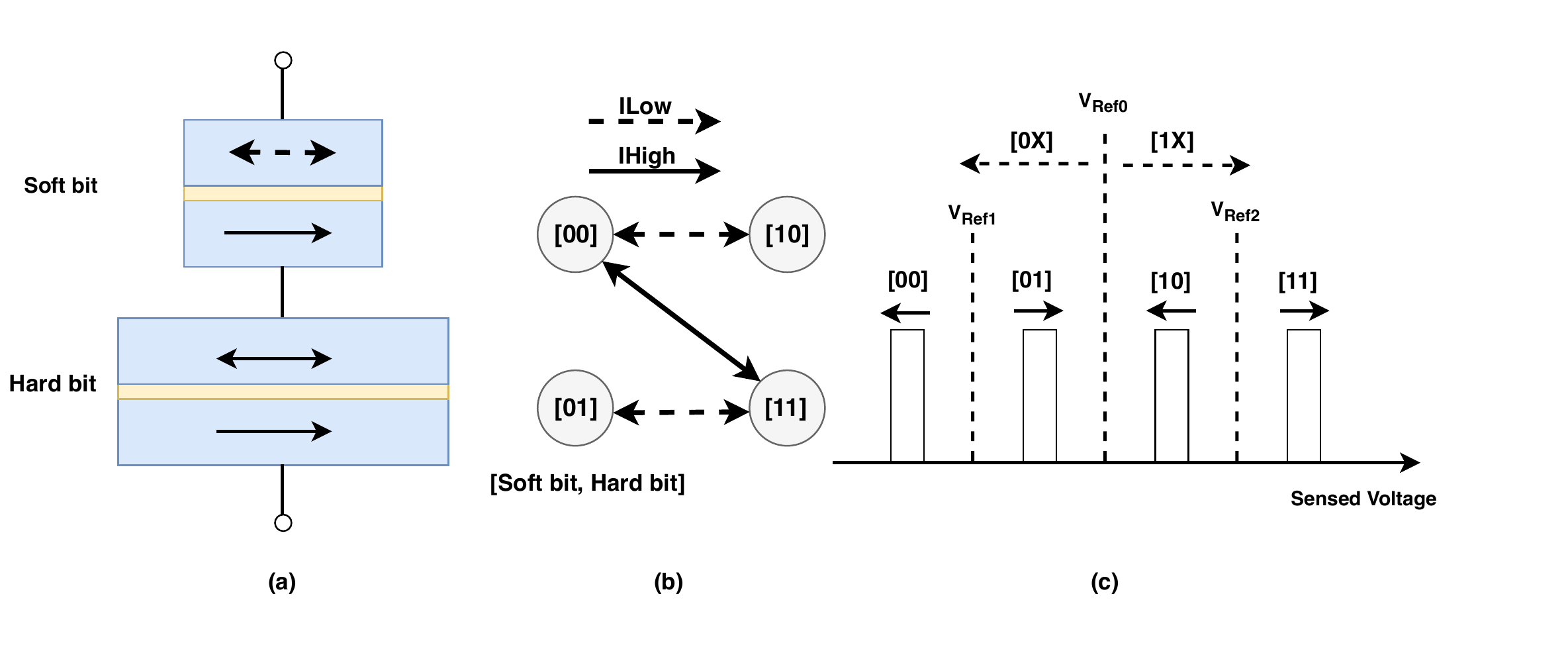}
  \label{fig:sub-second}

\caption{Multi-level cell STT-RAM.  (a) Series MLC MTJ structure; (b) 2-step write operation; (c) 2-step read operation.}
\label{fig:base}
\end{figure*}

Deep Neural Networks (DNNs) has become a very reliable solution for addressing many energy constraint problems over the last few years \cite{Energy,footprint}. Since large amount of data and computation is required for CNN operations, using proper hardware for this purpose is inevitable. Accelerators are energy-efficient devices that can carry out simple computation in a very effective manner. A typical accelerator-based architecture has been shown in Fig. \ref{fig:base}. In this system, a general-purpose architecture is connected to many Processing Elements (PE) through a shared medium. There is a DMA engine that can handle data transfers between main memory, CPU, and accelerators. In the accelerator side, there are 3 large buffers holding inputs, weights, and outputs. Buffers are responsible for keeping the PEs busy aiming for higher throughput. PEs are processing components that can compute simple functions such as add-multiply-sum operations. Eyeriss \cite{Eyeriss} categorizes the systolic arrays into five classes: Output Stationary (OS), Weight Stationary, (WS), Input Stationary (IS), Row Stationary (RS), and No Local Reuse (NLR). The difference is how tasks are assigned to computing nodes and how weights and inputs are distributed across the PEs. For each class weight matrices are differently mapped to a given MAC unit, and are not replaced until the computation is completed.
These design-choices have their own advantageous and disadvantageous, however, without loss of generality, we assume that our baseline is a weight-stationary system. The insight is that by keeping the weights more on PEs and not on an NVM buffer, the system would be more reliable.


\begin{figure}[t]

  \centering
   \includegraphics[width=3.4in]{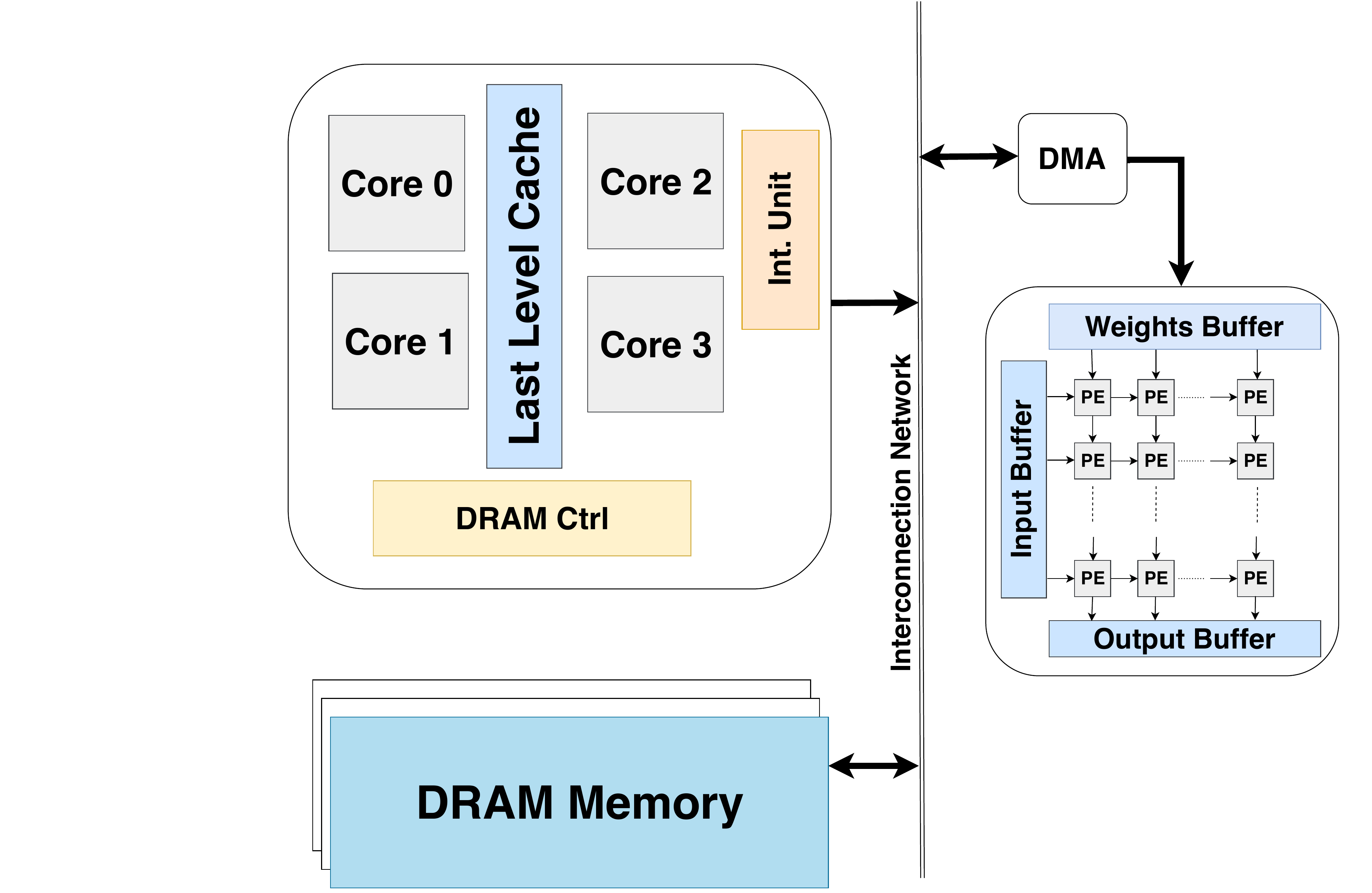}
\caption{Accelerator-based architecture for the CNNs operation}   
  \label{fig:sub-first}
  
\end{figure}

\subsection{MLC STT-RAM Basics}
MLC STT-RAM relies on a magnetic tunneling junction (MTJ) to store one bit. Basically, an STT-RAM cell has 2 layers, a free layer, and a reference layer. The reciprocal magnetic direction of these layers determines the value stored in each cell. The magnetization direction of the reference layer is fixed while the other one can change. To hold a "0" logic in a cell, a current must be applied from reference layer to free layer so that the magnetization direction of the free layer becomes the same as the fixed layer. This formation, which is called parallel has the lowest resistance and can represent logic "0". The magnetization direction of these two layers can be managed in such a way that resistance of a cell becomes very high representing logic "1". This design is called Single Level Cell (SLC) and is capable of storing one bit in each cell.

On the other hand, the structure can be extended in such a way that 2 or even more bits are stored in each cell. For this purpose, 2 MTJs are stacked to create 4 different and distinct resistances leading to a 2-bit cell configuration. One MTJ should be sized larger namely \emph{hard bit} and the smaller MTJ is called \emph{soft bit}. Fig. \ref{fig:base} shows the structure of the hard and soft bits. 

To program a 2-bit MLC STT-RAM, 2 steps are required. In the first step, the soft bit is programmed and then the hard bit is realized. More specifically, in the first step, we can program MTJ either to "00" or "11". Then in the next step, by applying another pulse, we can reach to "01" or "10". Fig. \ref{fig:base}(b) shows the program process of serial MLC STT-RAM. The solid line is the indicator of the first step and the dashed line shows the second step. For example, by applying a high current pulse to the STT cell, we can reach from "00" to "11". Then, by applying another pulse, we can work around the least significant bit. 

Read operation in a 2-bit MLC STT is performed as follows. First, a small current is applied to the cell and the resistance is compared to a reference value. Based on the result of comparison another pulse is applied and the result is then compared against another reference value. This approach is very similar to a binary search, where based on the comparison result, we narrow down the space of the search. As an example, let's say value "10" has been stored in the cell. To read the content, first, we apply a small current and compare either observed resistance or voltage to $V_{ref0}$. If the system is not faulty, the result of comparison leads us to apply the second pulse and compare the result to $V_{ref2}$. The results of this comparison tell us the voltage is lower than $V_{ref2}$ and higher than $V_{Ref0}$, and the stored value is realized.
The only consideration here is that the size of the current should be small enough not to change the value of a cell.


\subsection{Reliability of MLC STT-RAM Cells}
Process variations and thermal fluctuations in MTJ switching process are the two main sources of unreliability and inefficiency in MLC STT-RAM cells. Process variations persuade deviations of electrical and magnetic characteristics of MTJs from their nominal values and it causes read and write errors of STT-RAM \cite{staterestrict}. Furthermore, thermal fluctuations change the resistance switching process of MTJs, so that it causes uncertainty of switching time.

Write errors happen when the programming current is removed before the MTJ switching process is completed \cite{statisticalSTT}. In SLC cells raising the amplitude of programming current reduces the MTJ switching time and improves the write reliability \cite{ISCAS}. But in MLC cells, since the resistance difference between hard and soft bits is low, raising the amplitude of programming current in soft transition may cause flipping the resistance state of the large MTJ and overwriting the value held by the cell.


Sensing error and read disturbance are the two main sources of read operation failures in MLC STT-RAMs \cite{staterestrict}. Sensing error happens when MTJ resistance state can not be verified before ending the sensing period due to the small or false sense margin. In MLC STT-RAM the sense margin between adjacent states is smaller and therefore distinction between the resistance of states is harder than that of SLC.

Read disturbance occurs when the read current changes the resistance state of MTJ. This is also exacerbated with thermal fluctuations. In MLC STT-RAM cells since the probability of read disturbance is very low, it is ignored in most analysis \cite{staterestrict}.

\section{Related Work}
\label{sec:Related}
This section presents an overview of recent works on designing energy-efficient on-chip memory for CNN accelerators.
Most of the computations in CNNs/DNNs are based on matrix/vector multiplications and additions. There exist a considerable body of literature on performing these computations efficiently in hardware via GPUs \cite{GPU}, FPGA devices \cite{FPGA} and custom ASIC devices \cite{Diannao,Eyeriss,SCALEDEEP}.  
There have been studies to investigate memory footprint reduction through pruning between layers in neural networks \cite{DeepCompression,NIPS1992_647}. Some works reduced the precision of network's parameters to lower the number of required bits \cite{ReducedPrecisionSF}. However, works that reduce the precision policies can degrade the CNN accuracy. Authors in \cite{error_resilience} take advantage of error resiliency of machine learning application and tried to design energy-efficient accelerators. They employ a hybrid SLC/MLC memory to address the reliability issues of MLC system.  More clearly, some cells are written in MLC mode and the rest in SLC mode selectively with the aim of increasing the total reliability. The clear weakness of such a design is that effective capacity of memory system is reduced and the whole potential of MLC design is not unleashed. In our architecture, we do not sacrifice the capacity at all, and all cells operate in MLC mode.

NVM-based neural network accelerator has been the subject of research in \cite{PIM, PipeLayer,CELIA,DSE,MAXNVM}. These NVM-based accelerators usually focus on the fully connected layers to evaluate their idea. This is because a fully connected layer has much more weights than a convolutional layer and managing this amount of data is more important than convolutional layer weights. It must be noted that, in some works NVM is used as the logic not necessarily as the memory component.       

Some authors have focused on energy-efficient STT-RAM. They employed some techniques in both circuit and architecture levels \cite{STTCache,ReliableSTT,Cacherevive}. These works have fixed the high write energy of STT-RAM while conserving the accuracy of read and write operations. Authors in \cite{Stochastic} proposed embedding STT-RAM devices into neural networks as neurons. This work claimed that magnetic tunnel junctions can be interpreted as a stochastic memresistive synapse for neuromorphic systems. Another method employed by \cite{spintronic} proposed a quality-configurable single-level cell STT-RAM memory array. It stores data with different accuracy level based on the application requirement. All of mentioned techniques, are designed for special purposes and can not be used in a general neural network accelerators.

Authors in \cite{CNNBuffer} have applied a precision-tunable MLC STT-RAM buffer for energy-efficient general-purpose neural network accelerators. This work leverages error resilience feature of neural networks to tolerate the reliability issues in MLC STT-RAMs. In this work 16-bit fixed-point number system is used for representing data and weights. Our works is built on top of this work and further improves the reliability.

\section{Motivation}
\label{sec:Motivation}
There are 2 motivations behind this work. The first motivation is that limited range of weights leads to a situation where all covered numbers in the IEEE half-precision floating-point representation are not used. This leaves us some unused bits in the representation to be used as backup for other cell. By carefully deciding what bits to backup, we can improve the reliability of the system. The second observation is that MLC STT programming process is asymmetric. Two bit patterns "11" and "00" require less power to program while patterns "10" and "01" are energy-hungry. This feature can be exploited to enhance the energy efficiency of the system by increasing the number of "11" and "00" through data manipulations. In the following subsections, we first investigate these two observations with more details and then propose two schemes to exploit them. 

\subsection{Limited Range of Weights}
Many previous works noticed that weights in CNNs span in a short ranges \cite{WeightNormalization}. That has been the insight behind many pruning and quantization schemes \cite{DeepCompression,NIPS1992_647,Systematic}. According to these works weights are limited between -1 and 1 \cite{WeightNormalization}, because after any convolutional layers a weight normalization is performed. Having this observation in mind, we show that the second bit of the numbers is never be used. For better understanding, we show this phenomena through an example. 

Fig. \ref{fig:Float} shows 4 special numbers: "-1.0", "+1.0", "+1.99", and "+2.0" in full-precision floating-point representation. The first two rows in the figure represents "-1.0", and "1.0"; the largest numbers required by CNN. The first bit indicates the sign; negative in the first row and positive in the second row. Then in the exponent region, all bits are selected, but the second bit. These 2 cases show the biggest numbers that can be obtained when the second bit is unused. Fortunately, if we do not use the second bit, we can successfully cover any number between -1 and 1 because the largest numbers are already covered. Also, to cover any number between -1 and 1, we need to either reduce the exponent, or increase the mantissa, which in both case the second bit remains untouched. 

If the same exponent as the two previous ones is used (e.g. "01111111"), and if a non-zero value for the mantissa is picked, the number would have a value greater than 1 or less than -1, as it is a case in the third row of the example. So, we can conclude that for any number between -1 and 1, the second bit will not be used, and it can be a good candidate to be borrowed to host the sign bit. 
The very first number that utilizes the second is "+2.0", as shown in the last row, which is not used in CNNs anyway. More clearly, when the second bit is one, the exponent value is $2^7$=128 that needs to be subtracted from the bias which is 127. While the bits in mantissa part are all zero, the mantissa value would be 1.0. Hence, $1.0 \times 2^{128-127}=2$.

Note that while the second bit is not used, the first bit is used frequently. Because there is roughly even number of negative and positive numbers in the weights and parameters of CNNs. Therefore, we run into a situation where the pattern "10", and "00" happens a lot in the first two bits. If the number is positive, we have "00", which is reliable to be saved. However, if the number is negative, we get pattern "10", which is highly vulnerable to error and also needs a huge amount of power during the programming. Later, we show how we can utilize the second bit to store the weights safely. 
\begin{figure*}
      \centering
      \includegraphics[width=5in]{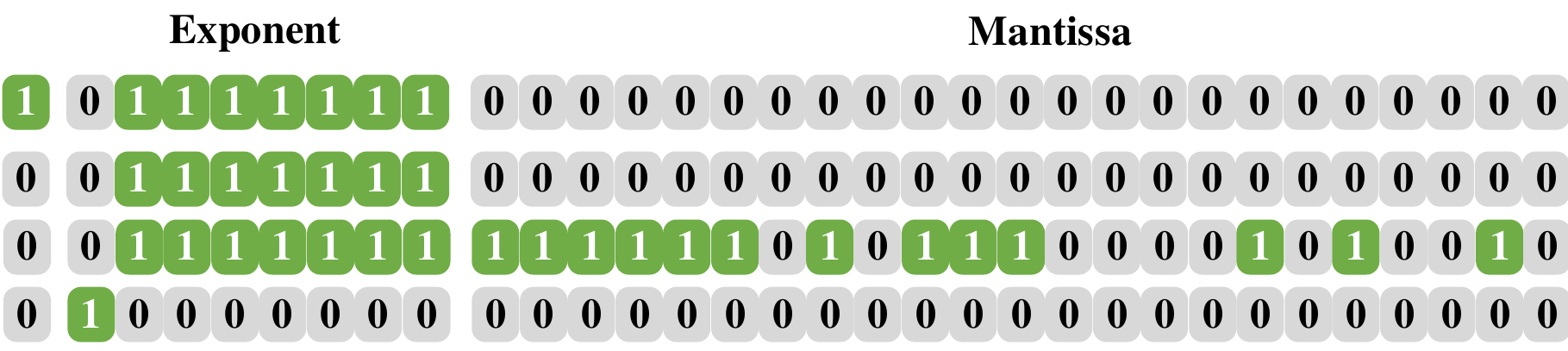}
      \caption{IEEE standard 754 floating-point representation. First bit is sign bit, the next 8 bits are exponents, and the rest are mantissa. There are 4 different numbers: -1, 1, 1.99, and 2 in full-precision floating-point representation.}
      \label{fig:Float}
\end{figure*}


\subsection{MLC STT Programming Asymmetry}
As mentioned in the previous section, when an MLC STT-RAM is programmed, patterns "00" and "11" need one iteration to finish, while patterns "10" and "01" require two iterations (Fig. \ref{fig:base}). Basically, in the first iteration, either the cell is programmed into "00" and "11", and then either the process is stopped or another step is taken to put the cell into "01" or "10". This way, we can claim that MLC STT-RAM programming is content-dependent. At 2-bit granularity, patterns "11" and "00" consume less power, and patterns "10" and "01" are slow and consume high power. Therefore the power consumption can be reduced, if any scheme can manipulate the data block in such a way that it gets fewer number of "01" and "10". 
Interestingly, the patterns "11" and "00" are also more resilient to soft errors. Because these 2 states are the base states and thus the cell has higher stability. In other words, patterns "11" and "00" are both power- and reliability-friendly.

In the next section, we show how by employing simple operations  the number of vulnerable bits can be reduced in the CNNs to better tolerate soft errors and also reduce the power consumption both at the same time.

\section{The Proposed Scheme}
\label{sec:proposed}

\subsection{Schemes}
We rely on the fact that the second most significant bit is always unused and can be utilized to save the sign bit (MSB) bit. Also, for STT-RAM, programming hard bit takes one iteration and soft bits takes up two iterations. Our goal is twofold; first protecting the MSB bit and second reforming the bit stream in such a way that number of "10" and "01" are reduced.

In this regard, we introduce three reformations as follows: 1) No Change, 2) Rotate Right by One, and 3) Rounding to Nearest. The first operation is called No Change because weights are written as is. The second operation is called Rotate Right by One, where the weight is rotated by one to the right. This can help when some patterns such as "10XXX01" appears in the bit stream and by rotating one bit, error-resilient patterns will place next to each other 110XXX0. Finally, the third operation tries to round the weight to the nearest MLC-friendly value. 

In order to figure out how many bits are required to be taken for rounding, we conduct an experiment where 1 million random numbers between -1 and 1 are generated. In this experiment, we flip one bit at a time and measure the error rate based on Error Sum of Squares (SSE). Fig. \ref{fig:Float} shows the SSE when different bit positions in half-precision floating-point are flipped. As can be seen from the figure, the last 4 bits have small impact and the SSE rate is very low. We limit the area where this scheme is applied to, because if we expand this area further (from last 4 digits to last 8 digits), the accuracy of classification will drop drastically. Based on our experiment to maintain the accuracy, it is best to select only the last 4 digits to be rounded. So, we take the last four bits and try to round them to an MLC-friendly value. To do so, since there are 4 MLC-friendly values ("0000", "0011", "1100", and "1111") in a 4-bit stream, we divide the 16 possible values into 4 classes uniformly and remap each to one the MLC-friendly value. The rounding process is shown in Tab.  \ref{tab:rounding}. As can be seen the first 4 values are assigned to "0000" and so on.

\begin{figure*}
      \centering
      \includegraphics[width=5in]{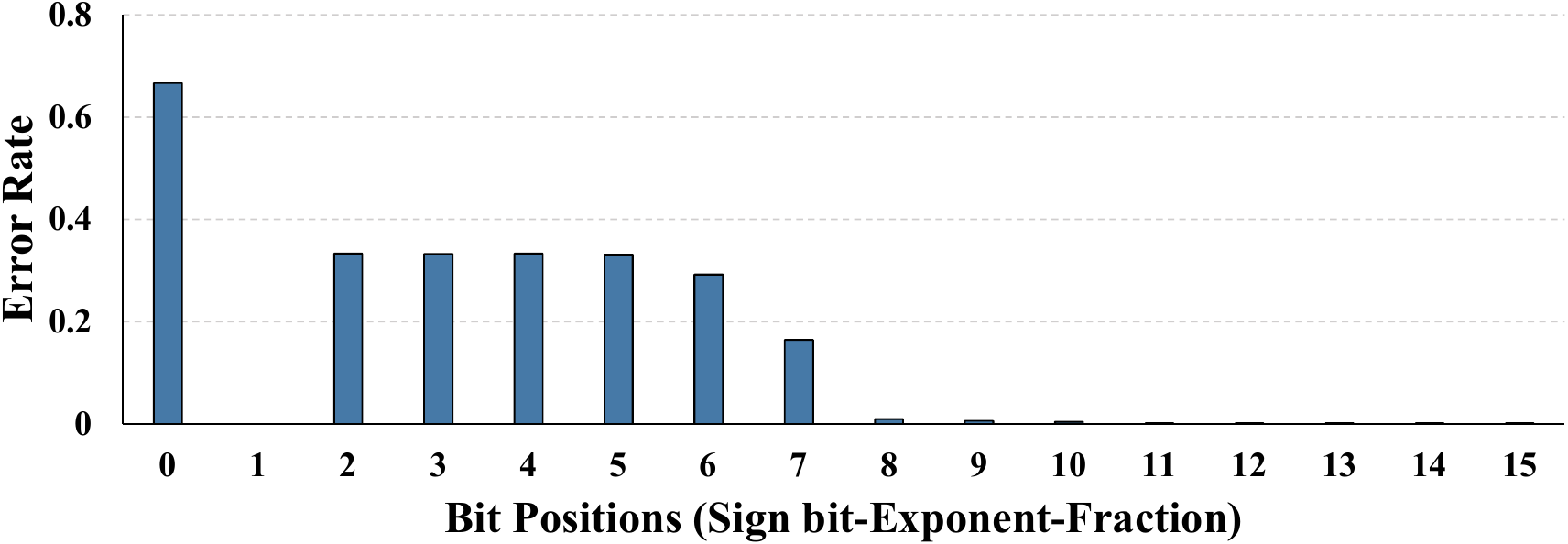}
      \caption{IEEE standard 754 floating-point representation. First bit is sign bit, the next 8 bits are exponents, and the rest are mantissa. There are 4 different numbers: -1, 1, 1.99, and 2 in full-precision floating-point representation.}
      \label{fig:Float}
\end{figure*}

\begin{table}[]
\caption{\label{tab:rounding} Rounding the bit patterns to MLC-friendly values.}
\begin{tabular}{|c|c|c|c||c|}
\hline
\multicolumn{4}{|c||}{Values} & Rounded \\ \hline
0000  & 0001  & 0010  & 0011 & 0000   \\ \hline\hline
0100  & 0101  & 0110  & 0111 & 0011    \\ \hline\hline
1000  & 1001  & 1010  & 1011 & 1100    \\ \hline\hline
1100  & 1101  & 1110  & 1111 & 1111    \\ \hline\hline
\end{tabular}
\end{table}

For better understanding, we explain each scheme by an example shown in Tab. \ref{tab:Example} and Fig. \ref{fig:Hybrid}. In our example we take three weights and try to reduce the number of soft bits. 

The first example in the Tab. \ref{tab:Example} is "0.004222". By looking at the binary representation in half-precision floating-point, we can see that patterns "00", "01", "10", and "11" occur 3, 3, 0, and 2 times, respectively. Then we sum up number of "11" and "00" after applying each three reformation schemes to compare with summation of "10" and "01". As can be seen from the last column in the table, we better to write this value unchanged as none of the other scheme is helpful. 

The next example in Tab. \ref{tab:Example} is 0.020614 where the binary representation has 2, 4, 1, and 1 bit patterns of "00","01", "10", and "11", respectively. However, if we rotate the bit-stream by one, as shown in the third row of the table, we can see that the number of soft bits is reduced from 5 to 3. Hence, in this situation, storing the weight in shifted format is the best option.

Finally, the last example in the Tab. \ref{tab:Example} is 0.0004982. As can be seen from the table, the number of "00" and "11" for No Change mode and Rotate mode are 4 and 4. Since CNNs are robust to inaccuracies, we round the last four digits to the nearest MLC friendly value based on the Tab. \ref{tab:rounding} mapping. For this particular example, since the last four digits are "0101", we round it to "0011". Doing so leads us to the situation where the number of soft bits is reduced to 2. 

\begin{table*}[]
\centering
\caption{\label{tab:Example} Examples for selection between 3 schemes (NoChange, Rotate, and Round).}
\begin{tabular}{|l|c||c|c|c|c|c|c|c|}
\hline
\textbf{Weight}        & \textbf{Binary}                          & \multicolumn{2}{c|}{\textbf{Operation}} & \textbf{00} & \textbf{01} & \textbf{10} & \textbf{11} & \textbf{Best} \\ \hline \hline
\multirow{3}{*}{0.004222} & \multirow{3}{*}{00 01 11 00 01 01 00 11} & NoChange    & 00 01 11 00 01 01 00 11   & 3           & 3           & 0           & 2           & \checkmark             \\ \cline{3-9} 
                       &                                          & Rotate      & 00 10 11 10 00 10 10 01   & 2           & 1           & 4           & 1           &              \\ \cline{3-9} 
                       &                                          & Round       & 00 01 11 00 01 01 00 00   & 4           & 3           & 0           & 1           &              \\ \hline \hline
\multirow{3}{*}{0.020614} & \multirow{3}{*}{00 10 01 01 01 00 01 11} & NoChange    & 00 10 01 01 01 00 01 11   & 2           & 4           & 1           & 1           &               \\ \cline{3-9} 
                       &                                          & Rotate      & 00 11 00 10 10 10 00 11   & 3           & 0           & 3           & 2           & \checkmark             \\ \cline{3-9} 
                       &                                          & Round       & 00 10 01 01 01 00 00 11   & 3           & 3           & 1           & 1           &               \\ \hline \hline
\multirow{3}{*}{0.0004982} & \multirow{3}{*}{00 01 00 00 00 01 01 01} & NoChange    & 00 01 00 00 00 01 01 01   & 4           & 4           & 0           & 0           &               \\ \cline{3-9} 
                       &                                          & Rotate      & 00 10 10 00 00 00 10 10   & 4           & 0           & 4           & 0           &               \\ \cline{3-9} 
                       &                                          & Round       & 00 01 00 00 00 01 00 11   & 5           & 2           & 0           & 1           & \checkmark             \\ \hline \hline
\end{tabular}
\end{table*}

\subsection{Overhead Analysis and Metadata}
We have to maintain some metadata to determine the mode of each weight (NoChange, Rotate, and Round). The metadata itself has to be stored in our STT-RAM memory, which is unreliable. Losing meta-data may cause a severe damage, because rotate may change the absolute value of the floating-point representation significantly. To overcome this difficulty, rather than having four schemes to utilize the 2-bit metadata, we proposed three schemes to store the metadata into a tri-level MLC, not a 2-bit MLC. As shown by many previous works, tri-level MLC is very reliable (close to SLC) \cite{staterestrict}. As a matter of fact, tri-level STT provides better error rate by sacrificing the information density; three states can be realized by tri-level STT and not by four. Using tri-level STT it is guaranteed that our metadata is safe and we will not impose any malfunction issues.  

From storage overhead point of view, we need to store 2 bits per each 16-bit weight leading to the overhead of 12.5\%. To reduce this overhead, we propose a grouping-based approach where weights are wrapped together and the best scheme is realized for each block of the wights. For example, we can apply our scheme at the granularity of four, where the three proposed schemes are examined for 4 weights together. Grouping weights together may slightly reduce the chance of finding the best scheme, however, can reduce the storage overhead significantly. 

\begin{figure*}[t]
      \centering
      \includegraphics[width=5in]{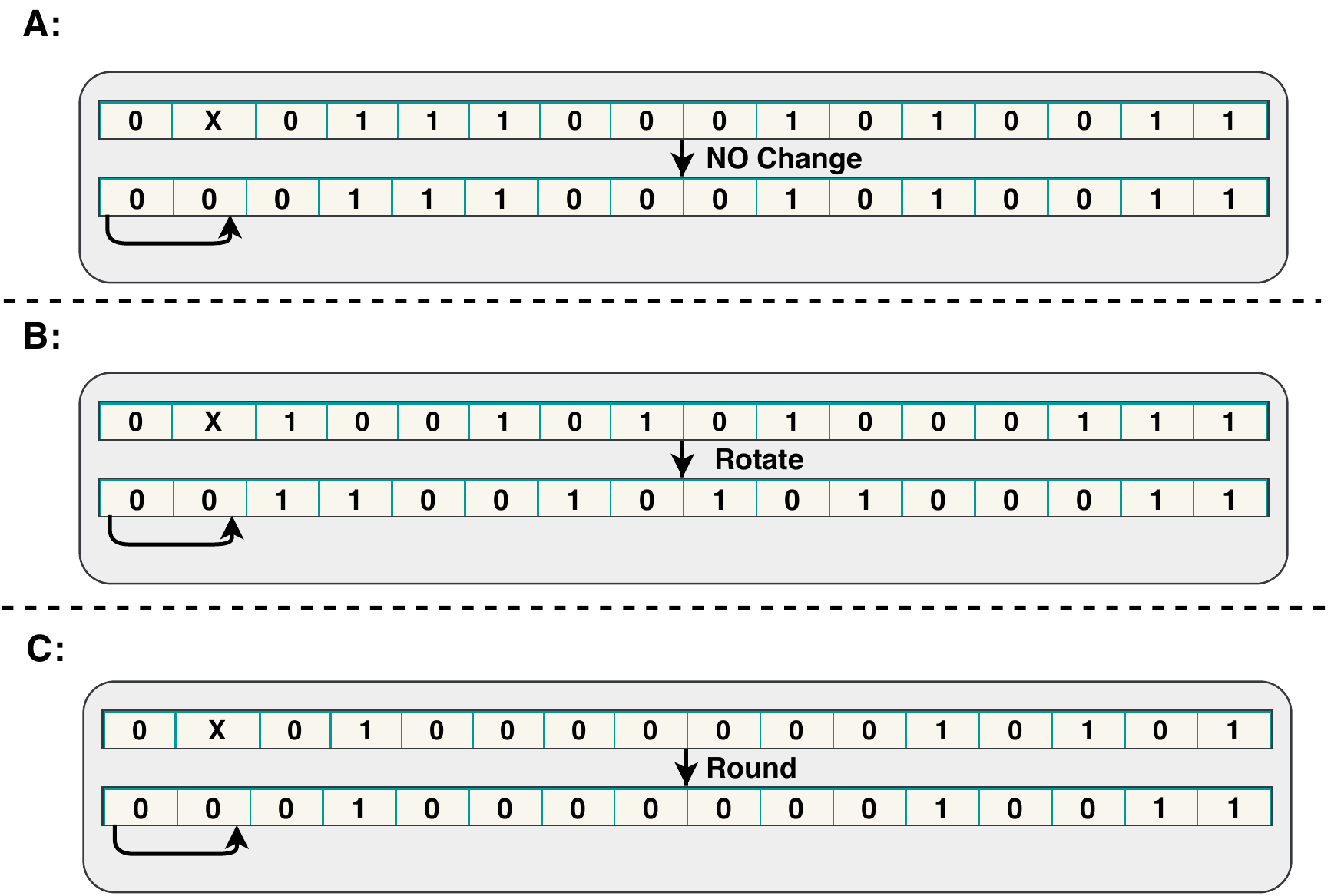}
      \caption{Final state of bit-stream for selecting the reliable weights}
      \label{fig:Hybrid}
\end{figure*}

\noindent{\textbf{Putting them all together }}
It must be noted that when we are rotating and rounding the weights, it is kept local. In other words, we apply to rotate scheme to each weight and count the soft and hard bits. At the same time, we count the soft bit when there is no change and when the rounding is applied to each weight. Then, we sum the number of soft bits together resulted from NoChanage/rotate/rounding to each weight separately. Finally, we chose the best scheme to be used as the final mode for the block.

Fig. \ref{fig:Hybrid} shows the final format of bit stream for the three examples shown in the Tab. \ref{tab:Example}. For the first case, only the sign bit is protected by duplication to the second bit. For the second case, in addition to protecting the sign bit, we rotate the bitstream by 1 to the right and then write the weight into the buffer. Finally, in the case of rounding, the last two cells are manged to store the nearest number.  Tab. \ref{tab:Overhead} shows the storage overhead for different granularity. As can be seen, the overhead can be reduced to less than 1\%.

\begin{table*}[]
\centering
\caption{\label{tab:Overhead} Storage overhead for different granularity}
\begin{tabular}{|l||l|}
\hline
Granularity & Overhead                                \\ \hline
1           & 2 bits/1 weight=16 bits  = 0.125        \\ \hline
2           & 2 bits/ 2 weights = 32 bits=0.0625      \\ \hline
4           & 2 bits / 4 weights = 64 bits=0.03125    \\ \hline
8           & 2 bits /8 weights= 128 bits=0.015625    \\ \hline
16          & 2 bits /16 weights = 256 bits=0.0078125 \\ \hline
\end{tabular}
\end{table*}

\section{Methodology}
\label{sec:Methodology}

\noindent \textbf{Classification Accuracy } We use Google Tensorflow \cite{TensorFlow} to evaluate two states of the art models: VGG16 and Inception V3. The input dataset to these models is ImageNet and we use transfer learning to train the models. The network is trained for 30 epochs, and the batch size of 100 is used.  

\noindent\textbf{Error model } In order to model the error induced by the MLC STT-RAM substrate, we use the previously proposed model \cite{compression}. In this model, read and write error rates are separated and faults are injected accordingly. To inject the errors, we assume all "00" and "11" are immune to soft errors because these two states are highly stable. However, for "01" and "10", we use a uniform fault injector to flip a bit. The probability of fault injections are taken from Ref. \cite{staterestrict} and are in the range of $1.5 \times 10 ^{-2}$ to $2 \times 10 ^{-2}$ \cite{staterestrict}. To incorporate the error model, we read all pre-trained weights and inject faults to the entire dataset. Then we store the pre-train weights to be used during the inference. We do not retrain the model because faults happen at the inference time by the memory substrate, and they will not be detected because of error detection complexity. So, it is not feasible to fine-tune the network after faults happen. Therefore, to be fair, we do not retrain the network. Finally, we report the accuracy of classification to judge between different systems. 

\noindent\textbf{Energy model} We use NVSim \cite{NVSIM} to evaluate the energy consumption of the proposed system. We also use the per bit energy cost reported in Tab. \ref{tab:Cost} to obtain the soft and hard bits cost of read and write operations. 

\noindent\textbf{Bandwidth Model} SCALESim \cite{SCALESim} is used to calculate the bandwidth of our systolic array. This simulator faithfully models a systolic array where all buffers are of the type of double-buffer.

\begin{table*}[t]
\centering
\caption{\label{tab:Cost} Soft and hard bits cost of reading and writing.}
\begin{tabular}{|l||c|c|l|}
\hline
\multicolumn{1}{|c||}{} & SLC STT-RAM & MLC STT-RAM & \multicolumn{1}{c|}{Hybrid} \\ \hline \hline
Read latency (cycle)   & 13          & 19          & Soft: 14, Hard: 20          \\ \hline
Write latency (cycle)  & 49          & 90          & Soft: 50, Hard: 95          \\ \hline
Read energy (nJ)       & 0.415       & 0.424       & Soft: 0.427, Hard: 0.579    \\ \hline
Write energy (nJ)      & 0.876       & 1.859       & Soft: 1.084, Hard: 2.653    \\ \hline
\end{tabular}
\end{table*}
\section{Evaluated Results}
\label{sec:Results}
In this section, we study the impact of our schemes on two models: VGG16 and Inception V3. Also, we show the results for 5 different granularity: 1, 2, 4, 8, and 16 words. 

\begin{figure*}
      \centering
      \includegraphics[width=5in]{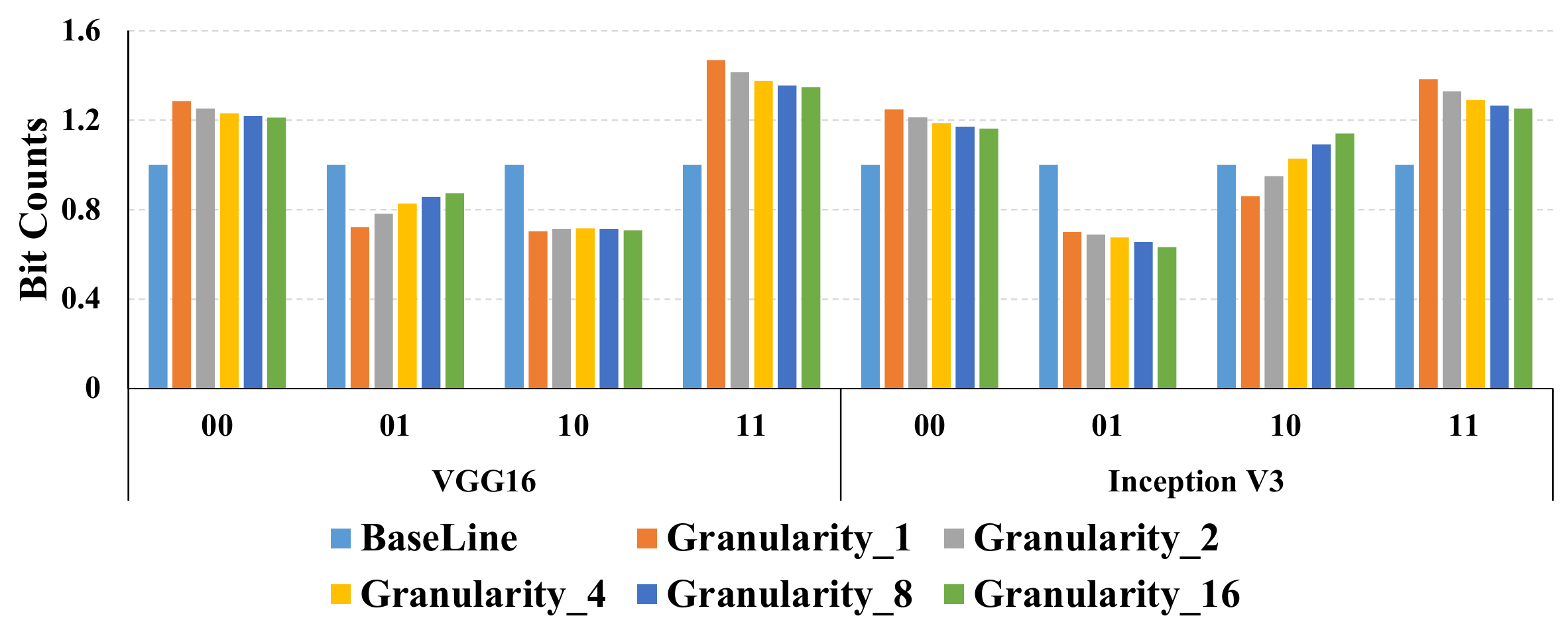}
      \caption{Bit count for 6 different systems, baseline and the proposed scheme with 5 different granularity.}
      \label{fig:BC_XX_Gran}
\end{figure*}

\noindent{\textbf{Bit count comparison }} Number of bit patterns have direct relation with power consumption and performance. In this experiment, we count how often different patterns are occurred. Fig. \ref{fig:BC_XX_Gran} shows the bit count for 6 different systems, baseline plus the proposed scheme with 5 different granularity. We show the results separately for VGG16 and Inception V3. As can be seen from the figure, Granularity\_1 shows a higher number of "00" and "11". As we increase the granularity, the number of "11" and "00" patterns decreases. However, the drop is not very significant, we only lose 5\% of these patterns if we increase the granularity from 1 to 16. Note that as the granularity increases, the storage overhead goes down as exhibited in Tab. \ref{tab:Overhead}. 

Fig. \ref{fig:BC_XX_Gran} shows that in VGG16 the "01" pattern increases as the granularity increases, while in Inception V3 we observe the opposite trend; pattern "01" stays the same, but pattern "10" increases.

\begin{figure*}
  \centering
   \includegraphics[height=1.4in]{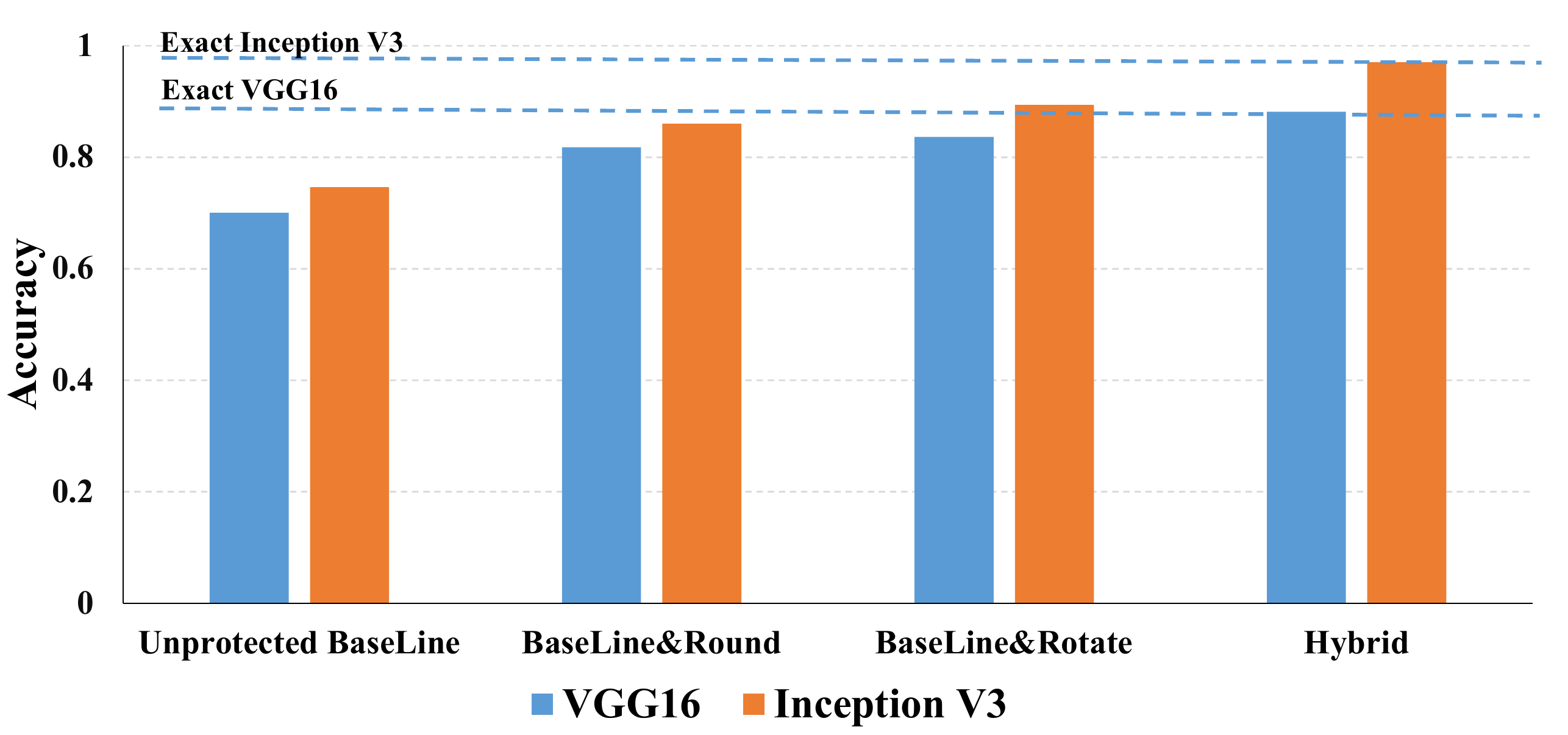}
  \label{fig:sub-second}
\caption{Accuracy for four different systems: 1) unprotected baseline, 2) Baseline+Rounding, 3) baseline+rotate and 4) baseline+rounding+rotate (hybrid).}
\label{fig:mot}
\end{figure*}

\noindent{\textbf{Energy Consumption }} Fig. \ref{fig:mot} (left), shows the energy consumption for different granularity and for read and write operations. Compared to the baseline all different granularity consume less energy. For example for Granularity\_1 of VGG16, the read energy consumption is reduced by 8\%, and for the largest granularity, the read energy is reduced by 7\%.  On the other hand, for Inception V3, the reduction of read energy is almost 8\%, while the write energy is reduced by 5\%. It must be noted similar to bit count results, when the granularity increases the gain degrades. This is due to the fact that fewer blocks are found to apply any scheme but NoChange, and the system is similar to the baseline.


\noindent\textbf{Classification Accuracy } Fig. \ref{fig:mot} (right) compares the accuracy for four different systems: 1) Unprotected Baseline, 2) Baseline+Rounding, 3) Baseline+Rotate and 4) Baseline+Rounding+Rotate (hybrid). Also, in this figure, the accuracy of both models in the error-free scenario are shown with dotted lines. When the system is unprotected (first bar) the classification accuracy significantly drops from 0.97 and 0.88 to 0.69 and 0.74, respectively. Now, we add our scheme one by one to the system to observe their impact. First, we  include the rounding in the second bar. When rounding is added to the baseline the accuracy increase 12\%, 11\%, respectively. Then, the rotate scheme is added to the baseline. As the result of including this scheme, the accuracy boosts up to 0.84 and 0.89. This scheme is slightly better than the rounding scheme independently.

Finally, the hybrid scheme is applied to the system. Hybrid here refers to a system where the best of (NoChange, Roatate and Rounding) is picked up. For the hybrid system, the classification accuracy reaches to the level of error-free scenario.  
Our hybrid scheme provides as good as accuracy compared to the error-free baseline. However, system 2 and 3 do slightly poorer than the error-free baseline. This figure shows that we can reduce the storage overhead further by applying only one scheme, but with lower accuracy. 


\begin{figure*}[t]

  \centering
     \includegraphics[width=5in]{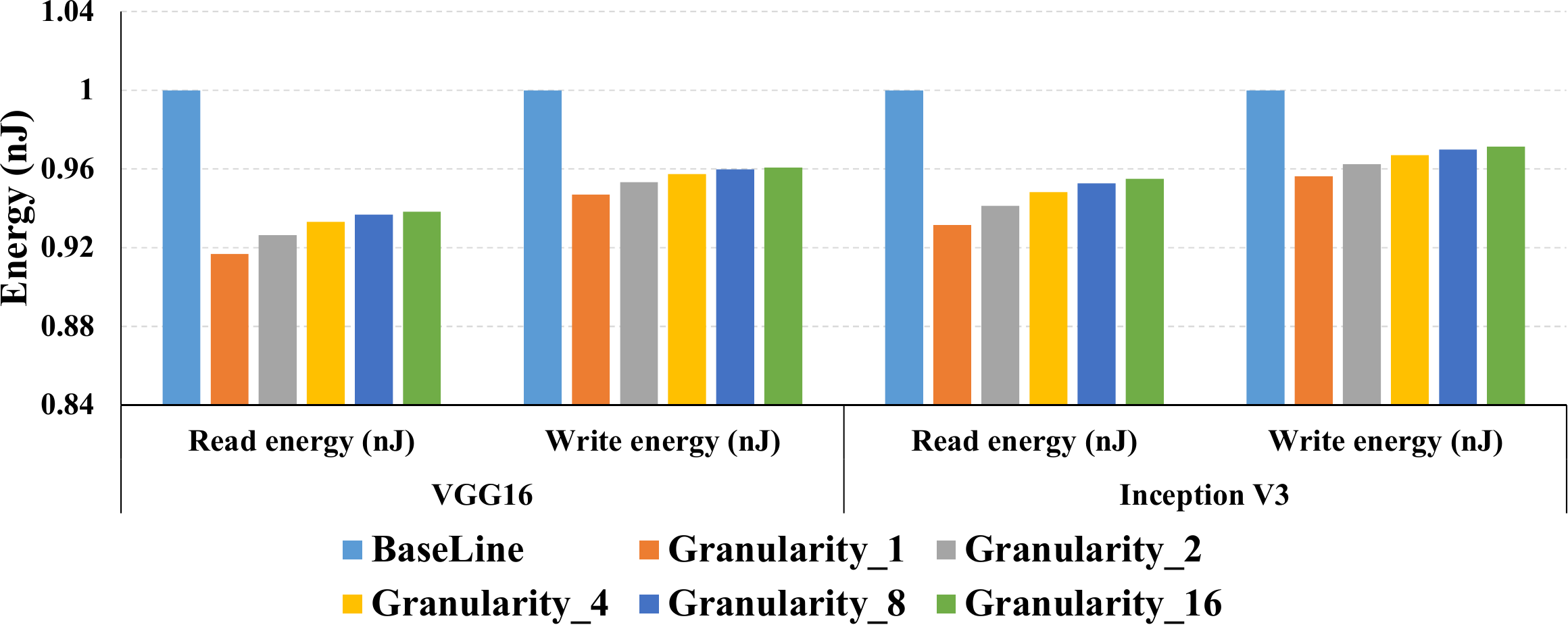}
  \label{fig:sub-first}
  \caption{Energy consumption for different granularity.}
\end{figure*}

\begin{figure*}
\begin{center}
      \includegraphics[width=\textwidth]{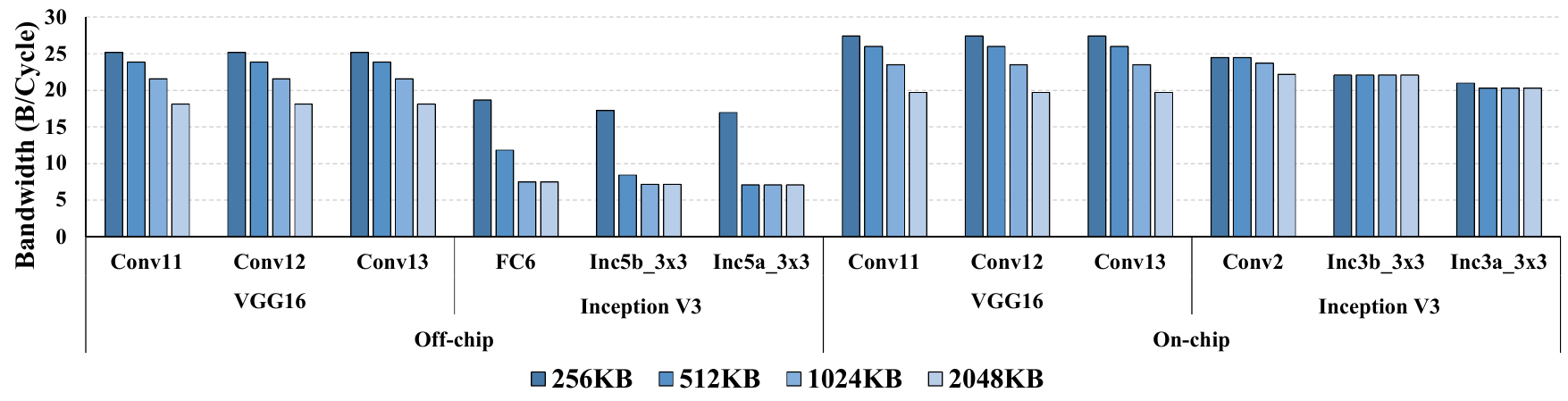}
\end{center}
      \caption{Maximum on-chip and off-chip bandwidth for different sizes of MLC STT-RAM.}
       \label{fig:BW}
\end{figure*}

\noindent \textbf{Bandwidth} Fig. \ref{fig:BW} demonstrates the bandwidth of memory sub-system for two cases: off-chip and on-chip traffics. Since there are many layers in each network, and there are 3 separate buffers in a systolic array, we report the maximum bandwidth for off- and on-chip traffic for top-3 layers in terms of bandwidth to account for the worst-case scenario. Also, the size of on-chip memory is varied from 256 KB to 2048 KB (ratio of 4). The system with 256 KB is an SRAM-based design, the rest are representative of a system equipped with MLC STT-RAM. For both cases (off-chip and on-chip), the required bandwidth is reduced significantly. For instance, In \emph{Conv11} layer of VGG16, the bandwidth is reduced from 25.5 bytes per cycle to roughly 17.1 bytes per cycle. The Inception V3 enjoys more from larger MLC STT-RAM buffers, and the required maximum bandwidth drops to 10 bytes per cycle with STT-RAM size of 2048 KB.

For the case of VGG16 on-chip traffic, the MLC STT-RAM is quite useful. The on-chip traffic is reduced by 24\% for \emph{Con12}. The on-chip bandwidth stays the same for two layers and slightly reduced for one layer in Inception V3. Although the on-chip traffic is larger than the  off-chip, but one can observe the advantages of the MLC STT-RAM. 

\section{Conclusion}
\label{sec:conclusion}
CNNs are becoming more and more popular due to their high accuracy and robustness. Newer models need larger memory to store their weights on-chip. To avoid the costly off-chip transactions, one solution is to increase memory capacity by employing emerging memory technology such as MLC STT-RAM. To address problems such as reliability and high dynamic power consummation associated with MLC STT-RAM, we propose a simple yet effective scheme to concurrently increase their reliability and reduce power consumption. Our hybrid scheme leverages from the fact that read and write operations are content-dependent, and thus data manipulation can impact the access time. In this regard, we devise a rounding and rotating mechanism to change the data block in such a way that the number of error-resilient patterns increases and at the same time the number of high-power patterns decreases. We chose the best option among the pure baseline, rotated, and rounding format solutions to achieve the highest level of reliability and accuracy. To overcome the difficulty of metadata management, we propose a grouping mechanism that combines some blocks together to further reduces the storage overhead. Our experimental results show that we can maintain the same level of accuracy as the baseline while reducing the read and write energy consumption by 9\% and 6\%, respectively.





\bibliographystyle{ieeetr}
\bibliography{ref}

\end{document}